\documentclass[12pt,a4paper]{cibb}

\PassOptionsToPackage{hyphens}{url}\usepackage{hyperref}

\makeatletter
\providecommand{\@ordinalM}[2]{#1}
\makeatother

\usepackage{subfigure,graphicx}
\usepackage{amsmath,amsfonts,latexsym,amssymb,euscript,xr}
\usepackage{booktabs}
\usepackage[nodayofweek]{datetime}
\usepackage{fmtcount}
\usepackage[english]{datenumber}
\usepackage[absolute]{textpos}

\usepackage[table]{xcolor}
\usepackage{color,colortbl,tabularx}

\usepackage[english]{babel}
\usepackage[protrusion=true,expansion=true]{microtype}
\usepackage{amsmath,amsfonts,amsthm}
\usepackage{pifont}

\definecolor{LightBlue}{rgb}{0.88,0.9,0.9}

\title{\Large $\ $\\ \bf Analysis of Prompt Engineering for Drug Toxicity Prediction}

\author{\large Mia MacGregor$^{*,1}$, Aakash Welgamage Don$^{2}$, and Mark Bartlett$^{1}$}
\address{\footnotesize $\ $\\$^1$ School of Computing, Engineering \& Technology, Robert Gordon University, Aberdeen, UK. \\
$^2$ School of Pharmacy, Applied Sciences \& Public Health, Robert Gordon University, Aberdeen, UK. \\
\bigskip
ORCID codes: FA 0009-0005-6021-2894; SA 0000-0001-8364-2495; TA 0000-0003-3383-4100.
\bigskip
\newline
$^*$corresponding author: m.macgregor1@rgu.ac.uk
\bigskip }

\abstract{\small Prompt Engineering, Machine Learning, Drug Toxicity. \normalsize
\\[17pt]
{\bf Abstract } Clinical trials in the UK can cost up to £1.3 million, with approximately 90\% drug failure rate. Toxicity is a major contributing factor in drug failure. Testing is time and cost intensive. In recent years, the use of artificial intelligence has been increasingly explored to aid in the prediction of drug toxicity, with extensive use of large language models (LLMs). However, LLMs can show considerable variation when minor changes are made to prompts, which raises concerns about their sensitivity to prompt engineering.  Prompt engineering is used to optimise a prompt given to an LLM to generate the desired output. This paper proposes a method to analyse prompt engineering for drug toxicity prediction.  The aim of the paper is to investigate the importance of prompt phrasing for drug toxicity prediction. LLMs were prompted to identify chemical properties of significance when predicting drug toxicity. Prompts were constructed to investigate; job role, prompt structuring, and rule interpretation. LLMs were then used to generate datasets, using the identified features from initial prompting, which were then passed to machine learning algorithms. The experiments show that the natural variance which occurs in LLMs outweighs any fine-tuning of prompts. There were, however, substantial improvements in model performance when using chemoinformatic code to extract features instead of using LLM-generated values. The proposed analysis methodology is applicable to a wide range of prompt types across different areas of bioinformatics. 
}

\abstract{\small Prompt Engineering, Machine Learning, Drug Toxicity. \normalsize
\\[17pt]
{\bf Abstract } 
Clinical trials in the UK can cost up to £1.3 million, with an approximately 90\% drug failure rate. Toxicity is a main contributing factor in drug failure. Traditional toxicity testing is time and cost intensive. In recent years, the use of artificial intelligence has been increasingly explored to aid in the prediction of drug toxicity, with extensive use of large language models (LLMs), such as ChatGPT. However, LLMs outputs can show considerable variation when minor changes are made to the prompts given to them. The aim of the paper is to investigate how important prompt phrasing is in relation to drug toxicity prediction. A machine learning pipeline was created that involved prompting LLMs at two points in the process. The experiments show that the natural variance which occurs in the outputs of LLMs outweighed any fine-tuning of prompts. There were, however, substantial improvements in model performance when using chemoinformatic code to extract features instead of using LLM generated values, which are common in the field. The proposed analysis methodology is applicable to a wide range of prompt types across different areas of bioinformatics. 
}
\usepackage{hyperref}

\begin{document}

\begingroup
\renewcommand{\thefootnote}{}
\footnotetext{\small{Article version: \datedate $\;$ h\currenttime $\;$ CET}}
\endgroup

\thispagestyle{myheadings}
\pagestyle{myheadings}
\markright{\tt Proceedings of CIBB 2026}

\section{Introduction}
\label{sec:SCIENTIFIC-BACKGROUND}
UK Clinical trials can cost up to £1.3 million \cite{mhra_fees_govuk} with only approximately 10\% of drugs included in phase I clinical trials reaching UK markets \cite{abpi_clinical_trials_2025}. One of the main reasons for failure in clinical trials is toxicity \cite{Sun2022DrugDevFailure}, which is a measure of harm that a chemical can cause to living organisms. Wet-lab toxicity prediction is associated with considerable time and cost burdens \cite{DennyStewart2024}.  

In recent years, scientists have begun using AI methods to predict toxicity prior to clinical trials, enhancing toxicity prediction by use of interdisciplinary tools, utilizing \emph{in silico} drug toxicity prediction \cite{Zhang2025-ot}. One method in recent literature is to use large language models (LLM) \cite{Song2025AdvancementsLLMs, Chung2025LLMToxicity, Yang2025-he, Zheng2025-xj}. The responses that LLMs provide can vastly differ depending on the prompt they are provided \cite{Song2025AdvancementsLLMs}, but the impact of the phrasing of the prompt used is often not considered in these studies. This paper addresses this gap by proposing a method to analyse prompt engineering used for drug toxicity prediction. Prompt engineering is the method used to optimise an LLM output \cite{HestonKhun2023}. We investigate the impact of prompt engineering on the variation of LLM outputs generated, as well as on the accuracy of the subsequent toxicity prediction task. 

Drug toxicity can be classified in different ways depending on the training data. Using different datasets allows our approach to investigate specific toxicity types, with toxic and non-toxic classifications determined by the machine learning (ML) models training data.


\subsection{Related and Background Work}
Recent years have seen growing interest in the application of AI for drug toxicity prediction. \cite{Zhang2025-ot}. 
Attention has recently shifted to using LLMs for this task. This section reviews some of the recent studies which adopt an approach similar to methodology studied in this work. 

\emph{Zero-shot methods} (such as \cite{Chung2025LLMToxicity}) ask an LLM whether a chemical is toxic without providing examples. \emph{Few-shot methods} ask the same question but also provide examples of the desired outputs for given inputs, as in \cite{Chung2025LLMToxicity,Yang2025-he}. These studies typically represent chemicals as SMILES (Simplified Molecular-Input Line-Entry System), which encode chemical information as a single line of text. \cite{Yang2025-he} combines a general prompt with a task-specific template, using GPT models in a few-shot setting by providing example SMILES with toxicity labels before asking for the toxicity of a SMILES with unknown toxicity. The LLM returns 0 (non-toxic) or 1 (toxic). However, it is unclear whether the model is predicting toxicity or recalling it from its training data (the \emph{data leakage} problem), making its ability to predict the toxicity of novel chemicals uncertain.


The state-of-the-art technique \cite{Zheng2025-xj} presents a methodology most similar to the one studied in this work. Rather than asking the LLM to directly classify the chemical as toxic or not, the LLM is asked to provide data about each chemical. This dataset is then used to train a ML model to predict toxicity; this avoids data leakage. LLMs undertake knowledge synthesis from the literature and then must identify inferred data rules from a labelled dataset to aid in toxicity prediction. These rules are used to create feature vectors for chemicals which contain information about each chemical stored as a list of 1's and 0's corresponding to a yes or no answer to a question. The vectors are then used to train ML models to predict the toxicity.

\section{Methods and Results}
\label{sec:DATA-AND-METHODS}
The general methodology used is described below, and each stage is explored in detail in subsequent sections. Initially, a prompt is passed to a LLM, asking it to output a list of 15 features that could be used in predicting the toxicity of a chemical. Unlike \cite{Zheng2025-xj}, we allow for features which are not just yes or no values, such as the inclusion of features like LogP or molecular weight. The LLM is then asked to produce the value associated with each feature for each of a list of SMILES obtained from PubChem \cite{PubChemBioAssay489025}. This results in a dataset containing a column for each of the 15 features and a row for each chemical. The ground truth for whether each chemical is toxic or not was also obtained from PubChem and appended to the dataset. The ML models were trained to predict toxicity using 80\% of the data and then evaluated on the remaining 20\%. The models produce feature importance values, and metrics are used to evaluate these. This process is repeated using different prompts and LLMs in order to study the impact of the prompt phrasing on the overall quality of the prediction made.

   
\subsection{Initial Prompt and Feature Generation}
\label{sec:init_prompt_dataset}

Prompts were designed to generate a list of 15 features that are of importance when predicting drug toxicity. Prompts were constructed to investigate job role, prompt structuring, and rule interpretation. These can be seen in Table~\ref{tab:Prompts}.

The experiments used several LLMs: Gemma3 4B \cite{Gemma2024OpenModels}, Deepseek-r1-Distill-Qwen-8B \cite{DeepSeekR12025}, Llama3.2 3B Instruct \cite{Touvron2023LLaMA}, Mistral 7B Instruct\cite{Jiang2023Mistral} and Gemini-2.5-flash \cite{DeepMind2023GeminiReport}. Gemini-2.5-flash was run via API using Google’s Gemini platform, while the other LLMs were run using Ollama\footnote{\url{https://ollama.com/}}. These LLMs were selected due to their occurrence in recent literature. Our intention was not to produce a state-of-the-art result but to allow us to investigate the variance between models when provided with the same prompts.

\begin{figure}[tb]
    \centering
    \includegraphics[width=0.75\linewidth]{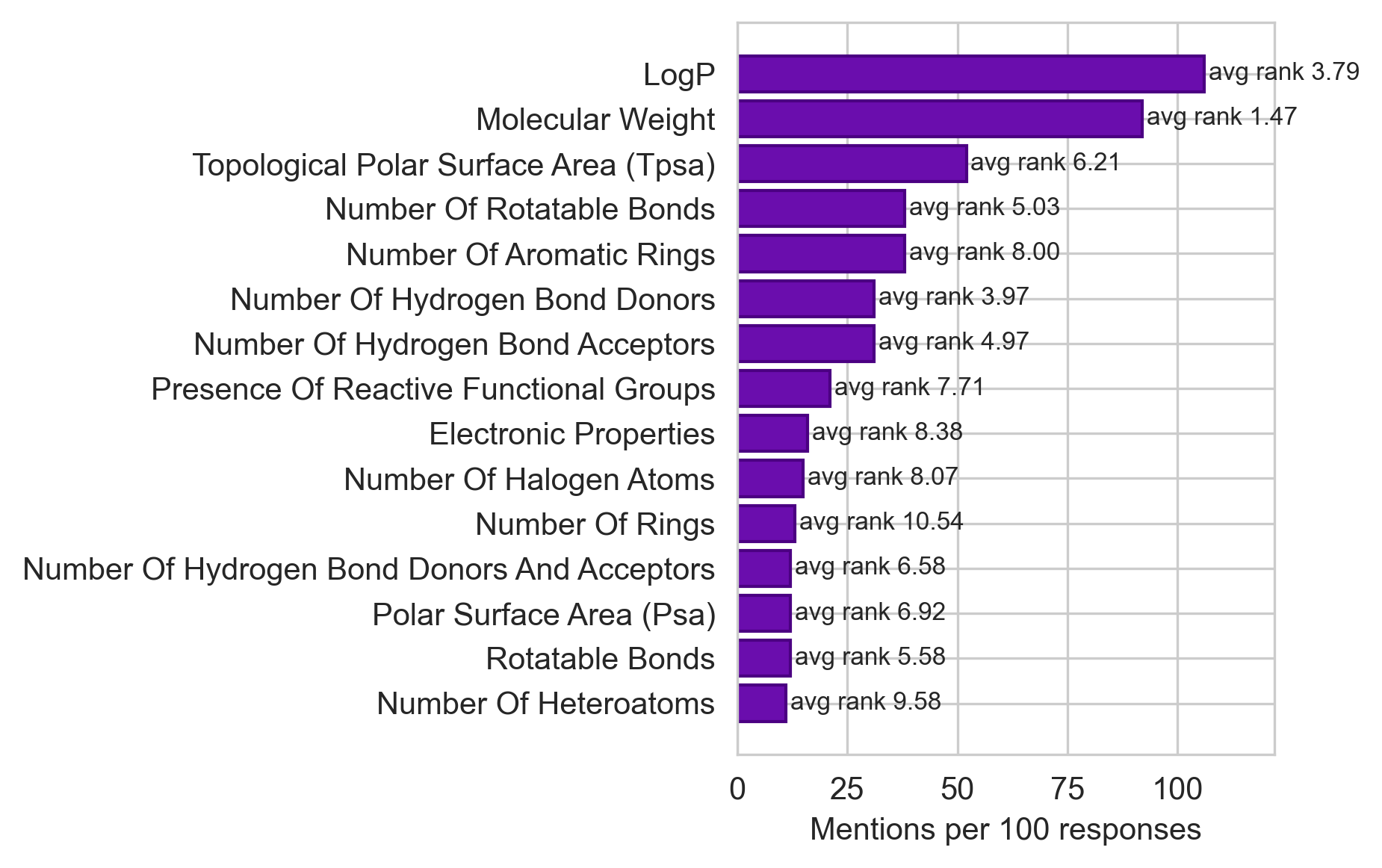}
    \caption{Most frequently identified chemical features across all prompt–LLM combinations.}
    \label{fig:TopFeaturesPer100}
\end{figure}

The output lists were analysed to identify the similarity of features generated from different LLMs and prompts. The feature names were standardised by Gemini 2.5 flash; this was necessary due to differing names being produced for the same property i.e. Molecular Weight, MolW, Mol Weight. The most recurring features can be seen in Figure~\ref{fig:TopFeaturesPer100}. LogP appeared in every feature list, this is expected as it is inversely correlated to water solubility, and can commonly indicate toxicity. Molecular weight and topological surface area were also prevalent in feature lists. There are only 7 features which occur in more than 25\% of the lists, showing a large degree of variance in most of the features identified by the LLMs.


\begin{table}[!tb]
\caption{Prompts passed to the LLMs. Each prompt was instantiated with four different job roles substituted for the \texttt{\{JOB\}} placeholder: Biochemist, Biologist, Chemist, and No job provided. This resulted in a total of 20 prompt variants tested. Colours indicate corresponding aspects of different prompts.} 
\label{tab:Prompts} 
\centering
\scriptsize

\begin{tabularx}{\textwidth}{|>{\raggedright\arraybackslash}p{2.5cm}|>{\raggedright\arraybackslash}X|}
\hline
\textbf{Prompt Variant} & \textbf{Prompt Text} \\
\hline
Prompt 1 – Baseline prompt & {\texttt{\{You are a \textbf{JOB}.\}}} Produce a list of fifteen items. \colorbox{green!20}{These should be factors which can be used to predict} \colorbox{green!20}{the toxicity of a chemical.} Do not produce any output except a numbered list.\\
\hline
Prompt 2 – Inclusion of SMILES &  {\texttt{\{You are a \textbf{JOB}.\}}} Produce a list of fifteen items.\colorbox{green!20}{These should be factors}\colorbox{yellow!20}{which can be calculated from} \colorbox{yellow!20}{SMILES}\colorbox{green!20}{to predict the toxicity of a chemical.} Do not produce any output except a numbered list.\\
\hline
Prompt 3 – Addition of contextual information &  {\texttt{\{You are a \textbf{JOB}.\}}} \colorbox{blue!20}{You are interested in toxicity and drug discovery. You want to know what features} \colorbox{blue!20}{would be of interest when investigating drug toxicity.} Produce a list of fifteen items. \colorbox{green!20}{These should be factors} \colorbox{green!20}{which can be used to predict the toxicity of a chemical.} Do not produce any output except a numbered list.\\
\hline
Prompt 4 – Combined SMILES features and contextual information &  {\texttt{\{You are a \textbf{JOB}.\}}} \colorbox{blue!20}{You are interested in toxicity of drugs. You want to know what features would be of} \colorbox{blue!20}{interest when investigating drug toxicity.} Produce a list of fifteen items.\colorbox{green!20}{These should be factors} \colorbox{yellow!20}{which can} \colorbox{yellow!20}{be calculated from SMILES}\colorbox{green!20}{to predict the toxicity of a chemical.} Do not produce any output except a numbered list.\\
\hline
Prompt 5 – Paraphrased prompt with role emphasis & {\texttt{\{You are a \textbf{JOB}.\}}}\colorbox{blue!20}{investigating drug toxicities. Which features would you look at to determine toxicity} \colorbox{blue!20}{of a chemical}\colorbox{yellow!20}{when only provided with a SMILES.}Produce a list of fifteen items. \colorbox{green!20}{These items should be} \colorbox{green!20}{used to predict chemical toxicity.}Do not produce any output except a numbered list.\\
\hline
\end{tabularx}
\end{table}

\subsection{Creation of Datasets}
To study the impact of LLMs, job role, prompt structure and inherent randomness on prediction accuracy, 18 datasets were generated using a subset of the previously obtained feature lists. The LLM was provided with 2,294 SMILES from a PubChem toxicity dataset \cite{PubChemBioAssay489025}. The dataset comprises 789 toxic and 1,505 non-toxic compounds, reflecting moderate class imbalance. The model also received a feature list from section~\ref{sec:init_prompt_dataset} and a prompt with instructions. Llama3.2 was selected due to its success in complying with strict formatting during preliminary experiments. Based on preliminary experiments, the below prompt was used for dataset construction.
\begin{footnotesize}
\begin{quote}
From the given SMILES string, compute all chemical features listed below.\\
Follow the exact feature order provided.\\
Output exactly 15 values, separated by commas only, no spaces.\\
Numeric values only. Binary features encoded as 1 or 0.\\
Output NA if a value cannot be computed.\\
If the SMILES is invalid, output INVALID\_SMILES.\\
Respond with the output line only.
\end{quote}
\end{footnotesize}

To check the accuracy of the LLM-generated values, we programmed a chemoinformatics pipeline to calculate the same features independently using RDKit. The pipeline used the same SMILES dataset and feature definitions, and produced results in the same format for comparison. Features that could not be calculated were either approximated using related descriptors or marked as missing. All calculations were deterministic. SMILES that could not be parsed were excluded from feature calculations but still recorded in the output.

Identical prompts were asked to the same LLM multiple times, to measure the impact of random variation (1aL\textsubscript{1}, 1aL\textsubscript{2} and 1aL\textsubscript{3}). We investigate using different LLMs, while keeping the prompt and role the same (3aD, 3aM, 3aL, 3aG and 3aF). Job roles were investigated by giving Llama3.2 the same prompt variant with differing job roles (5aL, 5bL, 5cL and 5dL). The datasets 4cL\_Calc and 4cL\_LLM used the same generated feature lists and compare extracting the features by chemoinformatic code or by an LLM respectively. A breakdown of the dataset names can be seen in the caption of Figure \ref{fig:AUCTop10heatmap}.




\subsection{Machine Learning Models and Evaluation}
The generated datasets were cleaned by removing uncomputable LLM-generated features and rows containing INVALID or NaN values. Ground truth toxicity was then added as a new column. The datasets were used to train several well-known ML models: Random Forest, Decision Tree, Neural Network (multilayer perceptron), Support Vector Machine, Naïve Bayes, and Extreme Gradient Boosting. StandardScaler, SMOTE, and GridSearchCV were used for scaling, data balancing, and parameter tuning, respectively.

\begin{figure}[tb]
    \centering
    \includegraphics[width=1\linewidth]{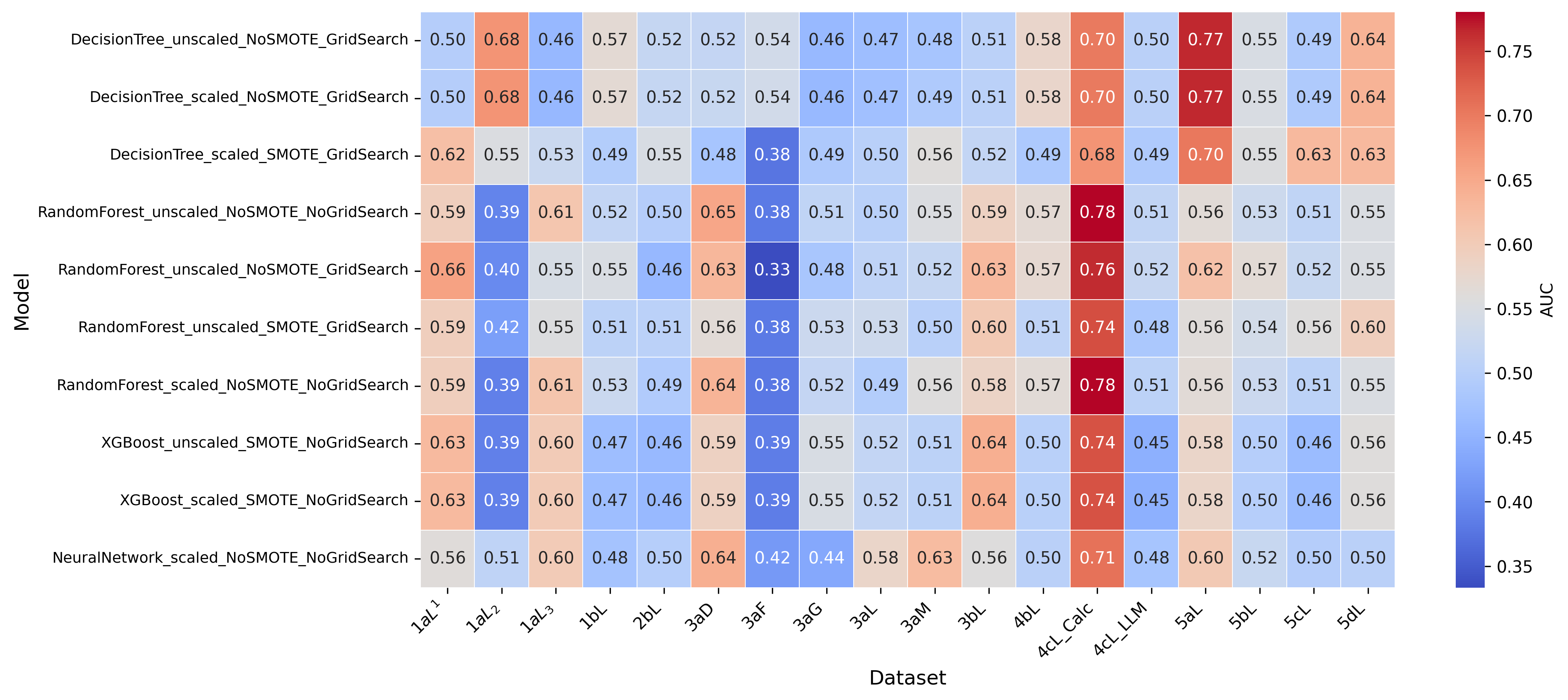}
    \caption {Heatmap of AUC values for the top 10 models across all datasets. Each cell shows the mean AUC of a model on a specific dataset. Warmer colours indicate higher AUC, highlighting models that perform consistently well across datasets. Dataset labels on the x-axis are a breakdown of the prompt and LLMs used. xyz - where x is the prompt used 1-5 as seen in Table~\ref{tab:Prompts}, y is the job variant; a=Biochemist, b=Biologist, c=Chemist, d=no role provided. z is the LLM used; D=Deepseek-v2, M=Mistral, L=Llama3.2, G=Gemma3 and F=Gemini-2.5-flash. Datasets which have xyzw when w is a number reflect repeated LLM sampling under inherent stochasticity in the generation process. If w =\_Calc this is where chemioinformatics code has been used to generate the dataset, when w =\_LLM this is the LLM-generated variant of the dataset.}
    \label{fig:AUCTop10heatmap}
\end{figure}

\begin{figure}[tb]
    \centering
    \includegraphics[width=1\linewidth]{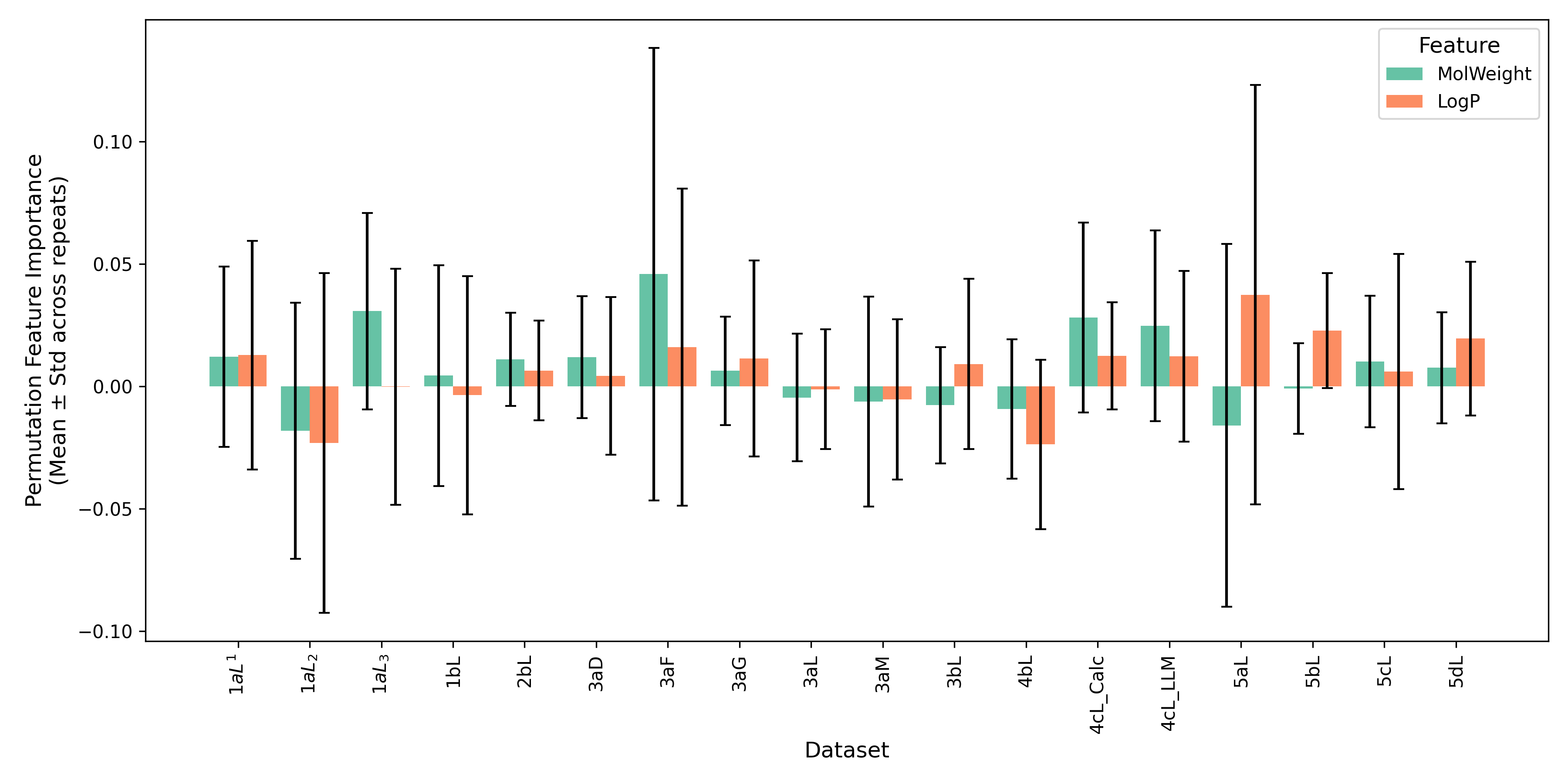}
    \caption{Permutation feature importance of Molecular Weight and LogP across all datasets and models. Bars represent mean decrease in model performance when a feature is permuted, calculated across multiple models. Error bars show standard deviation of feature importance between models. A breakdown in dataset titles can be seen in the caption of Figure.~\ref{fig:AUCTop10heatmap}.}
    \label{fig:MolWeightLogPimportance}
\end{figure}

 Figure.~\ref{fig:AUCTop10heatmap} shows that features extracted using the chemoinformatics pipeline outperform those generated by the LLM, as measured by AUC. Performance also varied across datasets when only the inherent randomness in LLM feature generation changed, indicating variability in the generated features. In contrast, changing the LLM had little impact, with similar AUC values across models. Performance also differed between job variants, with no single variant consistently performing best. All evaluation metrics showed similar trends. Permutation feature importance was calculated by disrupting a feature's relationship with the target and measuring the resulting drop in model performance. Feature importance for LogP and molecular weight across all models and datasets is shown in Figure.~\ref{fig:MolWeightLogPimportance}, illustrating the variation in the two most common features.

\section{Conclusions}
This study investigated the impact of prompt engineering on drug toxicity prediction. It was found that simple fine-tuning prompts did not have large effects on LLM output. The study found that random variation in LLM response seemed to cause similar variance to changing the prompt text or LLM used. Feature extraction by LLM was shown to be less effective than chemoinformatic extraction, with the latter showing increased performance. Although fine-tuning prompts may not be a productive approach in this instance to increase prediction quality, future work should investigate variations of the process, such as using representations other than SMILES, and more complex feature engineering methods such as retrieval-augmented generation. Significance testing may also provide additional insight into observed variations. 

\section*{Availability of data and software code }
\label{sec:AVAILABILITY}
The implementation is available at the following URL:  \footnote{\url{https://github.com/Mia-MacGregor/CIBB-Analysis-of-Prompt-Engineering-for-Drug-Toxicity-Prediction.git}}.


\footnotesize
\bibliographystyle{unsrt}
\bibliography{Ref_finalised.bib} 
\normalsize

\end{document}